\documentclass{article}

\usepackage{arxiv}

\usepackage[utf8]{inputenc} 
\usepackage[T1]{fontenc}    
\usepackage{hyperref}       
\usepackage{url}            
\usepackage{booktabs}       
\usepackage{amsfonts}       
\usepackage{nicefrac}       
\usepackage{microtype}      
\usepackage{lipsum}\usepackage{amsfonts}       
\usepackage{nicefrac}       
\usepackage{microtype}      
\usepackage{lipsum}
\usepackage{graphicx}
\usepackage{amsmath}
\usepackage{pbox}
\usepackage[usestackEOL]{stackengine}
\strutlongstacks{T}
\usepackage{longtable}
\usepackage{caption}
\usepackage[round]{natbib}
\usepackage{hyperref}
\title{A Survey on  Transfer Learning in Natural Language Processing}

\author{
  Zaid Alyafeai\\
  Department of Computer Science\\
  King Fahd University of Petroleum and Minerals\\
  Dhahran, Saudi Arabia 31261 \\
  \texttt{g201080740@kfupm.edu.sa} \\
   \And
 Maged Saeed AlShaibani \\
  Department of Computer Science\\
  King Fahd University of Petroleum and Minerals\\
  Dhahran, Saudi Arabia 31261 \\
  \texttt{g201381710@kfupm.edu.sa} \\
  \And
   Irfan Ahmad \\
  Department of Computer Science\\
  King Fahd University of Petroleum and Minerals\\
  Dhahran, Saudi Arabia 31261 \\
  \texttt{irfan.ahmad@kfupm.edu.sa} \\
}

\begin{document}
\maketitle

\begin{abstract}
Deep learning models usually require a huge amount of data. However, these large datasets are not always attainable. This is common in many challenging NLP tasks. Consider Neural Machine Translation, for instance, where curating such large datasets may not be possible specially for low resource languages. Another limitation of deep learning models is the demand for huge computing resources. These obstacles motivate research to question the possibility of knowledge transfer using large trained models. The demand for transfer learning is increasing as many large models are emerging. In this survey, we feature the recent transfer learning advances in the field of NLP. We also provide a taxonomy for categorizing different transfer learning approaches from the literature. 
\end{abstract}


\section{Introduction}
Humans have been communicating for thousands of years using natural languages. It is estimated that there are currently around 6,500 spoken languages around the world \citep{johnson2013what}. As the main method for communication, automating language understanding is a fundamental concept that has been studied for many years in the literature. As a result, many tasks, shown in Table 1, have been introduced to verify and validate these studies. 
Deep learning is employed to serve and craft models for these tasks as their complexity is, by an order of magnitude, out of the scope of the traditional machine learning algorithms. Training a deep learning model is not always affordable due to the huge computing resources and large datasets requirements for these models. This motivates to start exploring other directions to transfer knowledge from one deep learning model to another.

The general idea of transfer learning, formally defined in section \ref{sec:tl}, is to transfer parameters or knowledge from one trained model to another. Based on the availability of labeled dataset, transfer learning can be divided into transductive and inductive transfer learning (see section \ref{sec:tl}). These approaches are general and can be applied to many tasks in machine learning. For instance, \citep{pan2009survey} surveyed the literature for the transfer learning methods followed in recommendation systems with auxiliary data. Moreover,  \citep{xu2011survey} discussed multi tasking and transfer learning in the domain of bioinformatics using machine learning and data mining techniques. Recently transfer learning has been applied heavily in natural language processing. \citep{malte2019evolution} discussed the evolution of transfer learning in natural language processing. They mainly focus on the most dominant approach for transfer learning which is sequential fine tuning. However, we believe that transfer learning in NLP should be studied more thoroughly with highlights to all transfer learning approaches. 

In the past few years, language models have evolved and they achieved much better results compared to traditional language models. These trained language models were used to transfer knowledge to solve many natural language processing tasks.  In this survey, we highlight the latest advances, summarize them and categorize each one in the corresponding category of transfer learning. We follow a similar taxonomy developed by \citep{weiss2016survey} and \citep{pan2009survey} in categorizing and analyzing the literature. Since there is a huge number of papers related to natural language processing we focused only on the recent papers with a reasonable number of citations. 

\def\arraystretch{2}
\begin{table}[htp!]
\caption{Definitions of many natural language processing tasks.}
\begin{tabular}{|p{5cm}|p{10.5cm}|}
\hline
\textbf{Task}                  & \textbf{Description}                                                                                                               \\ \hline
Summarization             & The process of extracting the more important points in a set of documents and providing a shorter, more compact version      \\ \hline
Question Answering and Classification       & Given a question and a certain context, we want to index the answer to the question in the given context. This is usually called abstractive question answering. On the other hand, generative question answering considers the problem as a generative task. Another related task is question classification where we classify questions semantically to different categories.                    \\ \hline
Text Entailment           & Given a pair of sentences, we want to predict whether the truth of the first sentence implies the truth of the second sentence.               \\ \hline
Semantic Role Labeling    & A labeling task where each word is given its semantic role in the current context.                                           \\ \hline
Co-reference Resolution  & The process of collecting expressions that refer to the same object.                                                            \\ \hline
Named Entity Extraction and Recognition   & The task of extracting entities (extraction) along with their labels (recognition).                                                                     \\ \hline
Sentiment Analysis        & To classify sentences or paragraphs according to the sentiment, e.g., positive, negative or neutral.        \\ \hline
Reading Comprehension     & A similar problem to question answering where for a given context we want to comprehend it by answering questions correctly. \\ \hline
Translation               & The process of pairing each given sentence in a language $A$ to another sentence in language $B$ that has the same meaning.      \\ \hline
Sentence Pair Classification & Classify if a given pair of sentences are semantically equivalent.\\
\hline
Natural Language Understanding & Considers how effective models are on different tasks including  question answering, sentiment analysis,
and textual entailment, etc. \citep{wang2018glue}. 
\\
\hline
User Intent Classification & Association of text to a specific purpose or goal.\\
\hline
Natural Language Inference & Determine if given a premise whether the hypothesis is entailment, contradiction, or neutral. \\
\hline
Part of Speech Tagging & Label each word to its corresponding part of speech based on its meaning and context. \\
\hline 
Document Grounded Dialog Response & Given web document and conversation history what is the proper response ?\\
\hline
\end{tabular}
\end{table}

\section{Background}
\label{sec:background}
In this section we give a brief background of different models used for natural language processing. We divided the models into three categories based on the main architecture used . We also can think of recurrent-based models as being traditional because they have been recently replaced with parallelized architectures like attention-based and CNN-based models.

 \subsection{Recurrent-Based Models}
Recurrent neural networks (RNNs)  were first introduced as a way to process sequential data. The basic idea is learn the sequence context by passing the previous model state along with each input. RNNs showed good results in many tasks like time-series classification \citep{husken2003recurrent}, text generation \citep{sutskever2011generating}, biological modeling \citep{spoerer2017recurrent}, speech recognition \citep{graves2013speech}, translation \citep{bahdanau2014neural} and music classification \citep{choi2017convolutional}. Another variant of RNNs is the multi level hierarchical network introduced by \citep{schmidhuber.1992learning}. RNNs, unfortunately, suffers from an intrinsic problem. As many other machine learning algorithms, RNNs are optimized using back-propagation \citep{rumelhartet} and due to their sequential nature, the error decays severely as it travels back through the recurrent layers. This problem is known as the vanishing gradient problem \citep{hochreiter1998vanishing}.
Many ideas were introduced to recover RNNs from issue. One idea was to use Rectivited Linear Unit (ReLU) as a replacement for the Sigmoid function \citep{glorot2011deep}. Another idea  the introduction of Long Short Term Memory (LSTM) architecture \citep{gers1999learning}. The architecture is composed of multiple units with different numbers of gates, input, output and forget gates, in each unit. Each unit also outputs a state that can be used on the next input in the sequence. \cite{schuster1997bidirectional} introduces bidirectional LSTMs that can process sequences from forward and backward directions hoping that the network may develop better understanding from both sequence directions. Although this architecture can handle long sequence dependencies well, it has a clear disadvantage of being extremely slow due to the huge number of parameters to train. This leads to the development of Gated Recurrent Networks (GRUs) \citep{cho2014the} as a faster version of LSTMs. The reason why GRU are faster is that they only uses two gates, update and output. The authors, moreover, show that GRU architecture can be even beat LSTMs on some tasks such as automatic capturing the grammatical properties of the input sentences.  

\subsection{Attention-Based Models}

Recurrent neural networks suffer from the problem of being sequential and slow. Moreover, they can’t capture longer dependencies because of vanishing gradients \citep{hochreiter1998vanishing}. Moreover, RNNs treat each sequence of words as having the same weight with respect to the current processed word. Additionally, sequence activations are  aggregated in one vector which causes the learning process to forget about words that were fed in the past. The main idea of attention is allowing each word to attend differently to inputs based on a similarity score. Attention can be applied between different sequences or in the same sequence (self-attention). Generally attention can take many forms \citep{hu2019introductory} but in this section we focus on two main formulations:

\subsubsection{Soft Attention}

Soft attention general assigns weights to different inputs from a given sequence. Given a set of queries $Q$, keys $K$ and values $V$ we can evaluate an attention score using the following formula 

\begin{equation}
\text{Attention}(Q, K, V) = \text{softmax}(\frac{Qk^T}{\sqrt{d_k}})V
\end{equation}

The operation of attention can be broken out into three steps: 

\begin{enumerate}
    \item \textbf{Similarity}: this is calculated using the dot product which evaluates a similarity score between the two vectors $Q$ and $K$. Then the result is divided by a normalization factor which is a parameter related to the size of the model. 
    
    \item \textbf{Probability}: which is calculated using the softmax function. This calculates a probability score for each given sequence showing how much each word in the keys attends to a given word in the queries. Note, the set of queries and keys could be the same, we call this self-attention \citep{vaswani2017attention}. 
    
    \item \textbf{Weighting}: The weights calculated from the previous step is multiplied by the value matrix $V$. 
    
\end{enumerate}

This approach is typically called “Scaled Dot-Product Attention” as mentioned in the transformers paper \citep{vaswani2017attention}. There exists other types of soft attention like additive attention. In general, additive attention computes the attention scores using a feed-forward neural network \citep{bahdanau2014neural}. 

\subsubsection{Hard Attention}

Compared to soft attention, hard attention is considered a non-differentiable process. Hence, the gradients are estimated using a stochastic process and not gradient descent. Formally, given $s_t$ which indicates the position to attend to given the current word at position $t$.  We also define $s_{t,i}$ as one-hot encoding vector that sets to 1 at the $i$-the location if we want to attend to that position. We then can consider the attention locations as latent variables that can be modelled using a multinoulli distribution parameterized by $\alpha_i$ and $z_t$ as a random variable where, 

\begin{equation}
p(s_{t,i} = 1|s_{j<t},a) = \alpha_{t,i}
\end{equation}

\begin{equation}
    z_t = \sum_i{s_{t,i}a_i}
\end{equation}

We then can optimize the log-likelihood using a lower bound on the objective function \citep{xu2015show}.

\paragraph{Positional Encoding} In language modeling, we have a sequence of words with fixed positions for each word. However, the attention process we accounted for so far doesn’t apply any information about the position of each word. Compare this to RNNs where they by default apply a sequence of words, hence they inherently encode the positions of them.  On the other hand, in attention based models, encoded words can be treated out of order which might result in a randomization effect. One simple solution is to encode each word using some information about its position with respect to the current sequence.  After we embed each word using an embedding matrix, we extract the positions using the following formula:

\begin{equation}
    \text{PE}(\text{pos},2i)= \sin\left(\frac{\text{pos}}{10000^{2i/d_{model}}}\right)
\end{equation}
\begin{equation}
\text{PE}(\text{pos},2i+1)= \cos\left(\frac{\text{pos}}{10000^{2i/d_{model}}}\right)
\end{equation}

Given `$\text{pos}$` as the position within the sequence which takes the values $0 \cdots n-1$ and $i$ as the position within the embedding dimension which takes values $0 \cdots d_m-1$. 

\subsection{CNN-Based Models }
Convolutional neural networks (CNNs) were traditional proposed to be applied in image recognition tasks like character recognition \citep{lecun1998gradient}. The main ingredient of CNNs are convolutional and max-pooling layers for sub-sampling. The convolutional layers are responsible for extracting features and the pooling layers are used to reduced the spatial size of the extracted features. CNNs have been successfully applied in other fields in computer vision like image synthesis \citep{goodfellow2014generative}, object detection \citep{redmon2016you}, colorization \citep{zhang2016colorful}, etc. Although it is less intuitive, CNNs were also applied for natural language processing tasks. \citep{kim2014convolutional} applied CNNs for the task of sentence classification. Convolutional layers are applied on features extracted from word embeddings learned from Word2Vec embeddings \citep{mikolov2013efficient}. It was successful for many tasks like movie reviews, question classification, etc. Similarily, character-level CNNs were used for text classification \citep{zhang2015character}. CNNs where also combined with other architectures like LSTMs \citep{xu2015show} and attention \citep{you2016image}. Convolutional neural networks have been successful also in language modeling \citep{dauphin2017language}. They proposed a model with gated convolutional layers that can preserve larger contexts and can be parallelized compared to traditional recurrent neural networks. 
\section{Language Models}
\label{sec:lm}
Language modeling refers to the process of learning a probability distribution over set of tokens taken from a fixed vocabulary. Let $(x_1, x_2, \cdots , x_n)$ be a sequence of tokens, we want to learn a probability distribution of the form $P(x_1, x_2, \cdots , x_n)$ . The joint distribution is usually evaluated using the chain rule

\begin{equation}
    P(x) = \Pi_tP(x_t|x_{<t})
\end{equation}

Where $x<t$ represents all the tokens before the current token. Even though this is the most used technique there exists many other approaches in the literature:

\paragraph{Unidirectional LM} We consider only tokens that are to the left or right of the current context. For instance, given $(x_1, x_2, x_3, x_4)$, in order to predict $x_3$ we only use $(x_1, x_2)$ as a left context and $x_4$ as a right context. This can be applied in self attention by using triangular matrices such that the weights are zeros when we multiply by the current features. This is usually called auto-regressive encoding. 
 
\paragraph{Bidirectional LM} Every token can attend to any other token in the current context. For $x_3$ in  $(x_1, x_2, x_3, x_4)$ it sees the context of  $(x_1, x_2, x_4)$. The task of next word prediction becomes now trivial because any token can attend to the next word prediction. To prevent that, in the literature they usually use masked language models. 

\paragraph{Masked LM} Usually used in bidirectional LM where we randomly mask some tokens in the current context. The task is then to predict these masked tokens. The masked tokens are labeled as $\text{[MASK]}$. For instance, we will have the following representation $(x_1, x_2, \text{[MASK]} , x_4)$ and we want to predict the masked token as $x_3$. This is usually called denoising auto-encoding. 

\paragraph{Sequence-to-sequence LM} The input is usually split into two separate parts. Every token in the first part can see the context of any other token in that part. However, in the second part, every token can only attend to tokens to the left. For example, given the input $\text{[SOS]} x_1 x_2 \text{[EOS]} x_3 x_4 x_5 \text{[EOS]}$. Token $x_1$ can attend to any token from the first four while $x_4$ can only attend to the first 5.  

\paragraph{Permutation LM} This model attempts to combine the benefits of auto-regressive and auto-encoding. For a given sequence $X$ of size $T$ it can be factorized into $T!$ unique sequences. For instance, suppose we have a sequence  $(x_1, x_2, x_3, x_4)$ then a possible permutation will be of the form $(x_2, x_3, x_1, x_4)$. Then for a given token $x_1$ it can be conditioned on $x_3$ and $x_2$ only. Each permutation can be sampled randomly at each iteration step. 

\paragraph{Encoder-Decoder LM} While other approaches are usually implemented using a single stack of encoder/decoder blocks, in this approach both blocks are used. This task is powerful as it can be used for classification and generation \citep{raffel2019exploring}. Given a set of sequences $(x_1, x_2, x_3, x_4)$ encoded using an encoder we want to predict the next sequences $(x_2, x_3, x_4, x_5)$ using a decoder. 


\begin{table}[!htp]
\caption{A comparison between the different pretrained models in the literature}
\begin{tabular}{|p{3.5cm}|c|c|c|c|c|c|c|}
\hline
\textbf{Language Model} & \textbf{Transformer} & \textbf{ELMo} & \textbf{GPT-2} & \textbf{BERT} & \textbf{UNILM} & \textbf{T5} & \textbf{XLM} \\ \hline

Unidirectional & &   & \checkmark  &      & \checkmark    &    &     \\ \hline
Bidirectional   &     & \checkmark  &     & \checkmark   & \checkmark    & &     \\ \hline
Masked           &    &      &     & \checkmark    & \checkmark    &  \checkmark &     \\ \hline
Sequence to Sequence & &      &     & \checkmark    & \checkmark    &  &     \\ \hline
Permutation     &     &      &     &      &       &    & \checkmark  \\ \hline
Encoder-Decoder  & \checkmark &      &     &      &       &  \checkmark  &   \\ \hline
\end{tabular}
\end{table}

\section{Datasets}
There exists many datasets in the literature that are used for NLP tasks. In Table \ref{tab:datasets} we summarize some of the datasets used in the literature. Note that many datasets are not publicly available so they are not included. As noticed in the table \ref{tab:datasets}, the size of these datasets is quite large especially for those which came after 2012. This is due to the de facto deep learning requirement to train on such huge datasets.

The development of some of these datasets is improving over time. The first version of SQuAD, which can be considered as a solved task as some trained models, for instance \citep{lan2019albert}, can outperform humans. SQuAD 2.0 has 50K more questions compared to the previous version.

Other types of datasets provides a complete overview of many language understanding tasks, GLUE is a good example of such datasets where it comprises 9 tasks. Such types of datasets provide the model with more comprehensive insights about the language. Such dataset can be considered as a benchmark where many language models can compete to reach human level. In the next year, SuperGLUE is considered as an improvement where more challening tasks are included. 


\begin{table}[!htp]
\caption{Summary of different datasets used for transfer learning. }
\label{tab:datasets} 
\begin{tabular}{|p{4.5cm}|p{2cm}|p{4.5cm}|p{4cm}|}
\hline
\textbf{Reference} & \textbf{Dataset} & \textbf{Task} & \textbf{Size} \\ 
\hline
\citep{rajpurkar2018know} & SQuAD 2.0 & Question Answering & 150,000 questions \\
\hline
\citep{wang2018glue} & GLUE & General language Understanding & multiple tasks \\ 
\hline
\citep{wang2019superglue} & SuperGLUE & Challening  language Understanding & multiple tasks  \\ 
\hline
\citep{lai2017race} & RACE & Question Answering & 100,000 questions \\ 
\hline
\citep{maas2011learning} & IMDB & Sentiment Analysis & 50,000 reviews \\
\hline
\citep{sang2003introduction} & CoNLL -2003& Named Entity Recognition & 301,418 tokens for English \\
\hline
\citep{bowman2015large} & SNLI & Natural Language Inference & 570,000 pairs \\
\hline
\citep{johnson2016mimic} & MIMIC-III & patient notes deidentification (Named Entity Recognition) & 60,000 care unit admissions\\
\hline

\citep{stubbs2015automated} & i2b2-2016 & patient notes deidentification (Named Entity Recognition) & 41,142 \\
\hline

\end{tabular}
\end{table}

\section{Transfer Learning}
\label{sec:tl}
In this section we give an introduction to transfer learning. We mainly follow the same discussion and notations adopted in \citep{weiss2016survey} and \citep{pan2009survey} . We define a Domain $D$ as a tuple $(X, P(X))$ where $X$ is the feature space and $P(X)$ is the marginal probability of the feature space. We also define a task as a tuple $(y, P(y|x))$ where $y$ are the labels and $P(y|x)$ is the conditional distribution that we try to learn in our machine learning objective. As an example, consider the task of document classification. Then we can consider $X$ as the feature space of all the documents in the dataset. $P(X)$ is the distribution of the documents in the dataset. For each feature $x \in X$ we associate a label $y$ which is an integer. The task is associated with an objective function $P(y|x)$ which is optimized to learn the labels of the documents given the feature vector for each document. Given a source domain-task tuple $(D_s, T_s)$ and different target domain-task pair $(D_t, T_t)$, we define transfer learning as the process of using the source domain and task in the learning process of the target domain task. In Table \ref{tab:comb} we compare between different scenarios when the domain pair is different or the task pair is different. 
\begin{table}[!htp]
\caption{All possible combinations for the domain and task pair.}
\label{tab:comb}
\begin{tabular}[!htp]{|p{4cm}|p{11cm}|}
\hline
\textbf{Scenario}            & \textbf{Example in Sentiment classification}                                                                                                  \\ \hline
$X_s \neq X_t$             & The source domain could be English and the target could be Arabic.                                                                   \\ \hline
$P(X_s) \neq P(X_t)$       & The review could be written in the topic of hotels in the first domain while on restaurants on the target domain.                    \\ \hline
$y_s \neq y_t$             & As an example, the reviews in the source task might be binary while in the target task is categorical.                               \\ \hline
$P(y_s|x_s) \neq P(y_t|x_t)$ & For example, given a specific review in the domain task might have a label negative while in the target task it has a label neutral. \\ \hline
\end{tabular}
\end{table}
\subsection{Transductive Transfer Learning}
\begin{figure}
    \centering
    \includegraphics[scale=0.4]{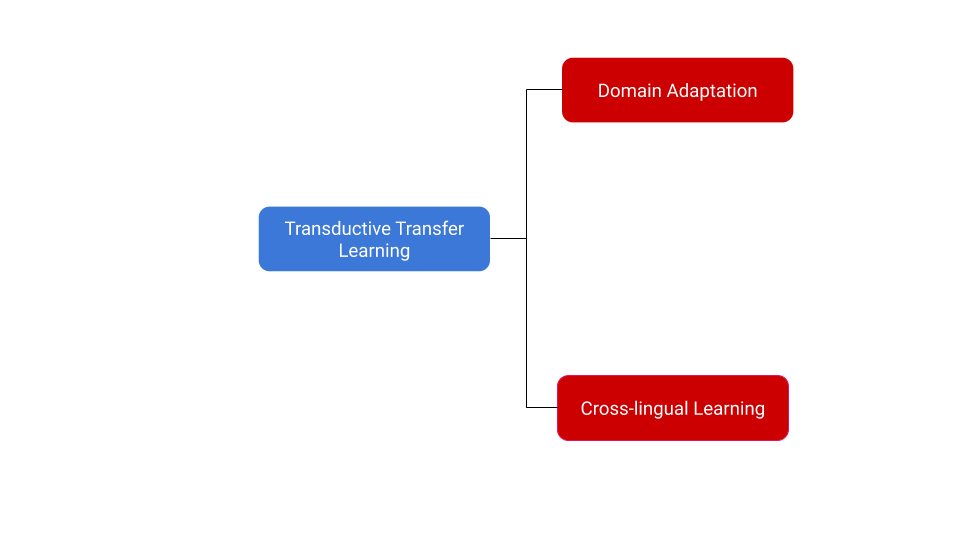}
    \caption{Transductive Transfer Learning}
    \label{fig:transductive}
\end{figure}

We have the same task where mostly we don't have labeled data in the target domain or we have a few labeled samples in the target task. Transductive transfer learning can be divided into two categories as shown in Figure \ref{fig:transductive}. 

\subsubsection{Domain Adaptation}

Refers to the process of adapting to a new domain. This usually happens when we want to learn a different data distribution in the target domain. For instance, in the field of sentiment classification, the review could be written on the topic of hotels in the first domain while on restaurants in the target domain. Domain adaptation is especially useful if the new task to train-on has a different distribution or the amount of labeled data is scarce. 

\subsubsection{Cross-lingual Learning}

Refers to the process of adapting to a different language in the target domain. This usually happens, when we want to use a high-resource language to learn corresponding tasks in a low-resource language. For instance, cross-lingual language modelling has been studied to see the effect on low-resource languages \citep{adams2017cross}.

\subsection{Inductive Transfer Learning}

We have different tasks in the source and the target domain where we have labeled data in the target domain only. Figure 2. Illustrates a taxonomy for dividing the different approaches that are mentioned in the literature. 

\subsubsection{Sequential Transfer Learning}

Refers to the process of learning multiple tasks in a sequential fashion. For instance, given a pretrained model $M$ we want to transfer the learning to multiple tasks $(T_1, T_2, \cdots , T_n)$. At each time step $t$ we learn a specific task $T_t$. Opposed to multi-task learning, it is slow but can be advantageous especially when not all the tasks are available at the time of training.   Sequential transfer learning can be further split into four categories.

\begin{enumerate}
 
    \item \textbf{Fine-tuning} given a pretrained model $M$ with weights $W$ for a new target task $T$ we will use $M$ to learn a new function $f$ that maps the parameters $f(W)=W'$. The parameters can be changed on all layers or on some layers. The learning rate could be different for the different layers (discriminative fine tuning).  For most of the tasks, we may add a new set of parameters $K$ such that $f(W, K) = W' \circ K'$.
    
    \item \textbf{Adapter modules} given a pretrained model $M$ with weights $W$ for a new target task $T$ we will initialize a new set of parameters that are less in magnitude than $W$, i.e., $K \ll W$. We assume that it can decompose $K$ and $W$ into smaller modules $K = \{k\}_n$ and $W = \{w\}_n$ reflecting the layers in the trained model $M$. Then we can define a function 
    $$f(K, W) =  k_1' \circ w_1 \circ \cdots k_n' \circ w_n $$
    Note that the set of original weights $W = \{w\}_n$ are kept unchanged during this process while the set of weights $K$ are  modified to $K' = \{k'\}_n$.
 
    
   \item \textbf{Feature based} only cares about learning some kind of representations on different levels like character, word, sentence or paragraph embeddings. The set of embeddings $E$ from a model $M$ are kept unchanged, i.e., $f(W, E) = E \circ W'$ where $W'$ is modified using fine tuning. 
    \begin{enumerate}
        \item \textbf{Character Embeddings} The characters are used for learning the embeddings. These models can be used to solve the open vocabulary problem \citep{ling2015finding}. 
        \item \textbf{Word Embeddings} The document is splitted by words and the words are encoded to create the embeddings. This is the most used approach with many techinques like Word2Vec \citep{mikolov2013efficient} and GloVe \citep{pennington2014glove}.
        \item \textbf{Sentence Embeddings} Sentences are used to create single vector representatiosn. For instance, the word vectors can be combined with N-grams to create sentence embeddings like Sent2Vec \citep{pagliardini2017unsupervised} 
    \end{enumerate}
   
    \item \textbf{Zero-shot} is the simplest approach across all the previous ones. Given a pretrained model $M$ with $W$ we make the assumptions that we cannot change the parameters $W$ or add new parameters $K$. In simple terms, we don't apply any training procedure to optimize/learn new parameters.
\end{enumerate}

\subsubsection{Multi-Tasks Learning}

Refers to the process of learning multiple tasks at the same time. For instance, given a pretrained model $M$, we want to transfer the learning to multiple tasks $(T_1, T_2, \cdots , T_n)$. All tasks are learnt in a parallel fashion. It can be argued that learning multiple tasks from a given dataset can be better than learning tasks independently \citep{evgeniou2004regularized}. 

 \begin{figure}
    \centering
    \includegraphics[scale=0.4]{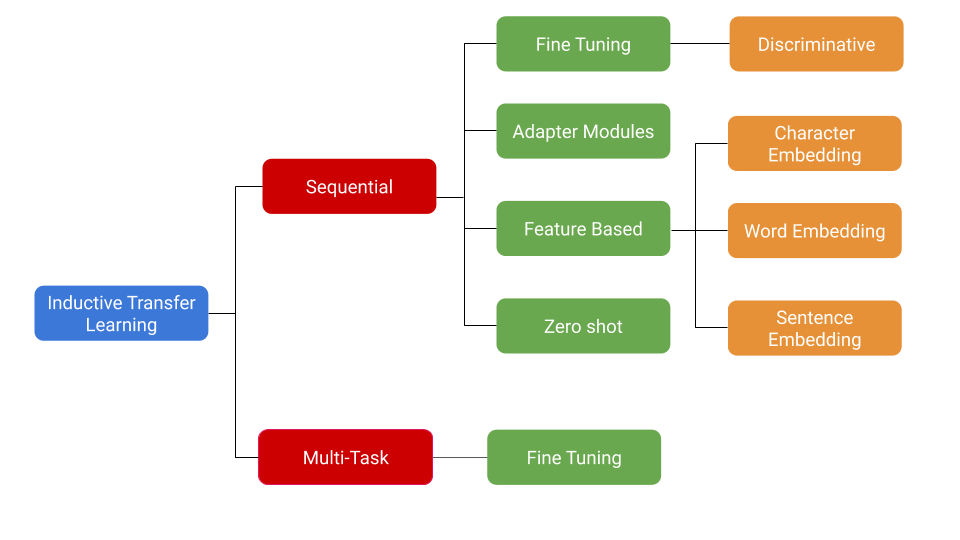}
    \caption{Inductive Transfer Learning}
    \label{fig:inductive}
\end{figure}

In the following section we provide a detailed literature review where we discuss the studies in each category. 

\section{Related Work}

There are many studies in the literature that discuss transfer learning in natural language processing. However, there are many overlaps in the mentioned studies and we did our best in order to divide them into their corresponding categories.

\subsection{Transductive Transfer Learning}
\subsubsection{Cross Lingual Transfer Learning}

\citep{kim2017cross} proposed a model for pos-tagging in a cross lingual setting where the input and output languages have different input sizes. The approach doesn't exploit any knowledge about the linguistic relation between the source and target languages. They create two bidirectional LSTMs (BLSTMs) one is called common where the parameters are shared between the languages and the private BLSTM where the parameters are language-specific. The outputs of the two modules are used to extract POS tagging using cross-entropy loss optimization. They force the common BLSTM to be language-agnostic using language adversarial training \citep{chen2018adversarial}. They show that this approach improves the results for POS tagging on 14 languages without any linguistic knowledge about the relation between the source and target languages. 

\citep{schuster2018cross-lingual}, presented a new dataset of 57k annotated utterances in English, Spanish, and Thai categorized in the following domains: weather, alarm, and reminder. This dataset is used to evaluate three different cross-lingual transfer methods on the task of user intent classification and time slot detection. These methods are: translating the training data, using cross-lingual pretrained embeddings, and novel methods of using multilingual machine translation encoders as contextual word representations. The results found on evaluating these approaches on the given dataset shows that the latter two methods outperform the translation method on a low resource target language, i.e., the target language has only several hundred training examples. This encourages investing more effort in building more sophisticated cross-lingual models.

\subsubsection{Domain Adaptation}
\citep{ruder2017knowledge} applied a teacher-student model for transferring knowledge from multiple domains to a single domain in an unsupervised approach. The student model is a multi-layer prerceptraon (MLP) that is trained in such way that it maximizes the similarity between the multiple source domains and the target domain. They use three measures for domain similarity which are Jensen-Shannon divergence \citep{remus2012domain},  Renyi divergence \citep{van2010using} and Maximum Mean Discrepancy \citep{tzeng2014deep}. They achieve state-of-the-art results on 8 out of 12 domain pairs for single source unsupervised domain adaptation. 

\citep{shah2018adversarial} applied adversarial domain adaptation for the detection of duplicate questions. Their approach consists of three components where the first component encodes the question. The encoder is optimized to fool the domain classifier that the question is form the target domain. The second component is a similarity function which calculates the probability that both questions are similar or duplicate. And a domain adaptation component which is responsible of decreasing the difference between the target and source domain distributions. The authors achieve an average improvement of around 5.6\% over the best benchmark for different pairs of domains. 

\citep{ma2019domain} considered the task of domain shift from pretrained BERT to other target domains. They use two steps where in the first step they classify the data points with similarity score to the source domain. Then, they use the classified data points to progressively train a model with increasing level of difficulty where lower difficulty means that the data point is similar to the target domain while higher means it is different. The later approach is called curriculum learning in the literature \citep{bengio2009curriculum}.  They tested their approach against four different large public datasets with varying domains and they outperformed other approaches on most transfer tasks in addition to faster training time. 

\subsection{Inductive Transfer Learning}

\subsubsection{Sequential Fine Tuning}

\citep{lee2017transfer} experimented the performance of transfer learning on the patient note de-identification task. This task is a variant of named entity recognition task where the model needs to find the patient's sensitive information in order to deidnetify. The transfer-learned model consists of a concatenated tokens embeddings and charachter embeddings feed to an LSTM unit. The output of this unit, then, goes through a fully connected layer for the seek of labels prediction. The output of this fully connected layer, labels vectors for each token, is finally fed to sequence optimization layer that outputs the most likely label of each token. The aforementioned model was trained on, initially, MIMIC dataset \citep{johnson2016mimic}, then, for the sake of fine tuning transfer learning, on i2b2 dataset \citep{stubbs2015automated}. They have conducted two experiments for two different goals. The goal of the first experiment was to quantify transferability by comparing the transfer learning approach against training the model on the target set only, the non transfer learning approach. While in the second one, they aimed to analyze the importance of each network layers' parameters when transfer learning on the target dataset. For the first experiment, they found out that transfer learning outperforms the non transfer learning approach on small train set sizes on the f1-score scale. This confirms the knowledge transferability. For the second experiment, the results show that transferring parameters of the bottom-most layers, mostly up to the LSTM unit, is as efficient as transferring the whole network. The work concludes that fine tuning the bottom-most layers, i.e. features layers, hold the most model knowledge. 

\citep{mccann2017learned}, tested the transferability of word vectors trained on machine translation datasets with different sizes. The trained model was an attention-based sequence-to-sequence model for English to German translation based on a research conducted by \citep{klein2017opennmt}. The architecture of this model, referred to as encoder, is a two layer bidirectional LSTM referred to as MT-LSTM. The resulting vectors out of this model are called Context Vectors, CoVe for short. Based on the size of the training dataset used, the vectors are of three variants, CoVe-S, CoVe-M, and CoVe-L. All of these datasets are tokenized using Moses Toolkit \citep{koehn2007moses}. The performance of this model and its variants was evaluated on two different NLP tasks: sentiment analysis, entailment and question classification and question answering tasks. The authors built a specific architecture using MT-LSTM for the purpose of classification. For question answering task, they used the Dynamic Coattention Network proposed by \citep{xiong2016dynamic} where the input is converted to CoVe vectors. The results produced by these models are competent to other models while some of these results outperforms the current state-of-the-art results in the literature. 

\citep{lakew2018transfer} proposed a method to transfer knowledge across neural machine translation models using a shared dynamic vocabulary. They have introduced two transfer learning approaches with different goals for each approach. The first approach, referred to as progressive adaptation (progAdapt) will dynamically update the embeddings of the target language based on the source language using an algorithm referred to as Dynamic Vocabulary update. The purpose of this approach is to maximize performance on the new target tasks from parameters learned from the previous tasks. The second approach, referred to as progressive growth (progGrow), is to initialize the translation model of the target language having the constraint that the performance of the source language model is maintained. This is achieved by feeding a language pair at a time to the model then updating the embeddings just as in progAdapt. The dataset used for the experiments was retrieved from WIT and TED corpus.

\citep{howardfine-tuned} proposed an approach to universal fine-tuning for text classification called ULMFiT. They follow a similar approach to the computer vision transfer learning task where a model is trained on ImageNet then fine-tuned to by adding classification layers at the end. They propose a system for universal fine-tuning that can achieve good results even on smaller datasets. They start by pretraining a AWD-LSTM model \citep{merity2017regularizing} on a large dataset. Then they fine-tune the model for a specific task by adding classification layers at the end. Then, they apply discriminative fine-tuning which works by applying different learning rates for different blocks in the pretrained model. The last layers in the model apply a higher learning rate than first layers. For training, they use slanted triangular learning rate (STLR), which increases linearly at the beginning then decreases at some point. Finally they apply gradual unfreezing which prevents catastrophic forgetting by unfreezing layers starting from the last layer. They show that their approach achieves state of the art on six text classification datasets. 

\citep{devlin2018bert} designed a model which is based on bidirectional encoder representation (BERT). They base their results on training BERT on large corpus of text then fine tune it by adding a small number of classification layers. This allows universal task fine tuning without substantial architecture modification. They base their results on learning bidirectional representations using a masked language model (MLM). To prevent bidirectional layers from attending to previous words, they mask out randomly some tokens and the objective is to predict the masked tokens. The authors argue that bidirectional representations are crucial for language understanding which makes it robust for different tasks like question answering. BERT architecture follows the transformer model \citep{vaswani2017attention}. To make BERT robust for different tasks they change the input representation to be a sentence-pair with some reserved token as a separator. Compared to traditional approaches, they train their models on two unsupervised tasks which are regular language model (LM) and next sentence prediction(NSP). For fine tuning, they simply plug the required inputs and outputs representations that are compatible for the tasks. For tasks that require only one input they use the empty string as the second pair. They provide empirical evidence that BERT achieves state-of-the-art results on eleven natural language tasks. 

\citep{peters2018deep} introduced deep contextualized word representations as word embeddings for language models (ELMo). They show that by applying deeper architectures they can achieve much better results than shallow words embeddings. Moreover, higher layers capture semantic relations while lower layers learn syntactic aspects like part of speech tagging. ELMo is based on bidirectional representations concatenated with the previous layer representation as a skip connection. For evaluation they test ELMo against six benchmark tasks including question answering, textual entailment, semantic role labeling, coreference resolution etc. They show that ELMo significantly improves performance on all six tasks. 

\citep{liu2019roberta} made some modifications to BERT to become robustly optimized (RoBERTa). The authors made a replication study to the BERT model \citep{devlin2018bert} with a few modifications. First, they trained the model on longer sequences over larger batch size and on more data. Second, they removed the next sentence prediction objective. Finally, they dynamically changed the masking pattern for each input sequence. They collected a large dataset of size 160GB (CC-News) which is comparable to other privately used datasets. They reached the conclusion that BERT was undertrained and with the previous modifications they can match or exceed other models published after BERT. They achieved state-of-the-art results on (question answering) SQuAD , (language understanding) GLUE and (reading comprehension) RACE. 

\citep{dai2019transformer-xl} introduced a new approach for preserving extra-long dependencies in language models (Transformer-XL). The main objective is to prevent context fragmentation which arises as a side-effect of training a language model on fixed size segments. Moreover, they used relative positional encoding instead of absolute positional encoding. In order to implement the idea, they follow the architecture of a Transformer \citep{vaswani2017attention} with the difference that later segments are conditioned on the previous segments. This allowed the model to learn dependencies up to 450\% more than vanilla transformers. Moreover, the evaluation speed is much faster than vanilla transformers. 

\citep{yang2019xlnet} suggested a modification to traditional language modeling by creating a permutation language model called XLNet. The approach focuses on taking the best of both  Autoregressive Language Modeling (AR) and Denoising Autoencoding (DA) while solving the main issue of two paradigms. The authors recognized that they should apply bidirectional encoding but without using masks like BERT because they want to limit the ability of such powerful language models. Mainly, they condition the current input on a permutation of all current tokens using masked attention. Moreover, to prevent two inputs from being conditioned on the same factorization they use target-aware representations where the representations are conditioned on the target positions as well. In order to preserve longer dependencies they use the architecture Transformer-XL. The authors show that XLNet outperforms BERT on 20 tasks by a large margin including question answering, natural language inference, sentiment analysis, and document ranking. 

\citep{dong2019unified} proposed a model for unified pretrained language model (UNILM). The main idea is combining different training objectives to pretrain a model in a unified way. They mainly combine four objectives: Unidirectional, Bidirectional and Sequence-to-Sequence (refer to section \ref{sec:background}). They evaluate the UNILM model on different tasks including abstractive summarization, generative question answering and document-grounded dialog response generation. They achieve state-of-the-art results on all of the previous tasks. 

\citep{roberts2020how} studied the knowledge retrieval performance of large language models. They investigated that in the domain of open-domain question answering with the constraint that we cannot look up any external resources to answer the questions. This task is similar to a closed-book exam where students are not allowed to look up books for answering exam questions. As a pretrained model they used the T5 model by  \citep{raffel2019exploring} which has 11 billion parameters. The hypothesis is that such a large language model with a huge number of parameters can store knowledge and hence we can extract this knowledge for a specific task. Moreover, T5 is a text-to-text model which makes it suitable for an open domain question answering task. Fore prediction, the token with the highest probability is decoded as the next prediction at a specific time step. They map this task to the T5 model by using the question as an input with the task-specific label and predict the answer as an output. They show that this approach outperforms models that explicitly look up answers using an external domain.

\subsubsection{Adapter Modules}

\citep{houlsby2019parameter-efficient} investigated how to make fine-tuning more parameter-efficient. In an online setting, tasks that arrive in stream can be trained efficiently with minimal parameters increase. These models are called compact models because they require a few number of parameters to fine tune. They typically call their approach “Adapter-based tuning” which can achieve 2x lesser parameters compared to traditional fine tuning with comparable accuracy. This allows training on different tasks in a sequential manner without the need for all the datasets for all tasks. The adapter based approach works by injecting layers in-between the layers of the pretrained model. The weights of the pretrained model are kept frozen, on the other hand, the new weights are initialized using a near-identity function. This prevented the model from failing to train. The adapbter based modules consist of feedforward layers, nonlinearity and a skip connection. They tested their approach by using BERT as a pre-trained model and 26 diverse text classification tasks.

\citep{stickland2019bert} applied adapter modules to share the parameters between different tasks by fine-tuning the BERT model. They propose projected attention layers (PALs) which are low dimensional multi-head attention layers that are trained with the attention layers of BERT in parallel. They also add task-specific residual adapter modules \citep{rebuffi2018efficient} within BERT to jointly learn multiple tasks. They evaluated their models against GLUE tasks while obtaining state-of-the-art results on text entailment. \citep{semnani2019bert} fine-tuned Bert on the task of question answering which they call BERT-A. They applied a combination of adapter modules \citep{houlsby2019parameter-efficient} and projected attention layers \citep{stickland2019bert} in order to reduce the cost of retraining the BERT model. They applied answer pointers where the start token is fed into a GRU in order to predict the end token \citep{wang2016machine}. They were able to reduce the number of parameters to 0.57\% compared to retraining all the parameters of BERT while achieving comparable results to state-of-the-art results on question answering.  

\subsubsection{Feature Based }

\citep{mou2016how} studied the transferability of models on similar tasks. They studied two general transferring methods: initializing models with parameters trained on other models (INIT) and training the same architecture on different tasks (MULT). The first experiment was an LSTM-RNN to classify sentences according to their sentiment. The datasets used in this experiment are IMDB, MR, and QC. IMDB and MR are for sentiment classification while QC is a small dataset for 6-way question classification. Transfer learning occurs between IMDB $\to$ MR and IMDB $\to$ QC with variant types of transferring methods, i.e., word embeddings are initialized using Word2Vec, randomly initialized, transferred but fine-tuned and transferred but frozen. The paper reports accuracy enhancement for each task with no more than 6\%.
The second experiment was a sentence pair classification. The model was a CNN-pair from Siamese architecture. The dataset used is SNLI \citep{bowman2015large}, a large dataset for sentence entailment recognition, SICK a small dataset with the same purpose as SNLI, and MSRP a small dataset for paraphrase detection. The transfer learning task was applied on SNLI $\to$ SICK and SNLI $\to$ MSRP. Improvements in the results was around 6\%. For layer transfer analysis (MULT), the paper showed a slight improvement for similar tasks but not for semantically different tasks. Combining both methods was not promising, according to their conclusion. 

\citep{peters2017semi-supervised}, utilized a semi supervised transfer learning approach for the task of sequence labeling. The model they used is a pre-trained neural language model that was trained in an unsupervised approach. The model is a bidirectional language model where both forward and backward hidden states are concatenated together. The output of this model is, then, augmented to token representations and fed to a supervised sequence tagging model (TagLM). The sequence tagging model is then trained in a supervised way to output the tag of each sequence.  The dataset used to conduct the experiments was CoNLL 2003 NER and CoNLL 200 chunking. They achieve state-of-the-art results on both tasks compared to other forms of transfer learning.

\subsubsection{Zero-shot}

\citep{radford2019language} proposed a large model for unsupervised multi-task learning. They train a set of large models ranging from 117 million parameters to 1.5 billion parameters which they call (GPT-2). They show that given a large model trained on large corpus can achieve zero-shot knowledge transfer without any fine-tuning. For such a task they created a dataset of size 40 GB text. For the language model, they use a similar architecture to the transformer model \citep{vaswani2017attention} and  GPT model \citep{radford2018improving}. The performance improvement follows a log-linear fashion as the capacity of the model increases. The largest model (GPT-2), achieves state-of-the-art results on 7 out of 8 different datasets. These tasks include summarization, reading comprehension, translation, and question answering. More recently \citep{gpt-3} was proposed with the largest model that has 175 billion parameters. The authors show that bigger models with bigger datasets (45TB of compressed plain-text) can achieve near state of the art results in a zero-shot setting. GPT-3 has a similar architecture to GPT-2 with the exception that they use "alternating dense and locally banded sparse
attention patterns". 

\citep{yin2019benchmarking} studied zero-shot transfer on text classification. They first modeled each classification task as a text entailment problem where the positive class means there is an entailment and negative class means there is non. Then they used a pretrained Bert model on text classification in a zero-shot scenario to classify texts in different tasks like topic categorization, emotion detection, and situation frame detection. They compare there approach against unsupervised tasks like Word2Vec \citep{mikolov2013efficient} and they achieve better results in two out of the three tasks. 

\subsubsection{Multi-task Fine Tuning}

\citep{liu2019multi-task} studied multi task deep neural networks (MT-DNN). The training procedure takes two stages. The first stage uses a BERT model to train a language understanding model. The second stage contains adding 4 models for task specific adaptation. These tasks include Single-Sentence Classification, Text Similarity, Pairwise Text Classification and Reverlance Ranking. For each task they added a task-specific objective that is minimized jointly during the training procedure. They evaluated the MT-DNN model on different GLUE tasks. They also compared applying the MT-DNN without fine tuning and with fine tuning and they found out that with fine-tuning they achieve the best results. They show that compared to BERT, large MT-DNN achieves state-of-the-art results on GLUE tasks.

\citep{raffel2019exploring} explored the effect of using unified text-to-text transfer transformer (T5). They follow a similar architecture to the Transformers model \citep{vaswani2017attention} with an encoder-decoder network. However, they used fully-visible masking instead of casual masking especially for inputs that require predictions based on a prefix like translation. Basically, they remove the necessity of masking for some tasks. To train the models, they created a dataset that is extracted from the common crawl dataset. Namely, they produced around 750GB dataset by cleaning a huge number of web documents. To train such a large dataset they used a huge number of parameters; hence the largest model contains around 11 billion parameters. To perform well on different tasks, they used multi-task pretrained models where the models were explicitly trained on different tasks by using prefixes like: “Translate English to German”. By fine tuning on different tasks like summarization, text classification, question answering, etc. they achieve state-of-the-art results on these tasks.

\begin{longtable}[!htp]{|p{4cm}|p{2cm}|p{4cm}|p{4cm}|}
\caption{Summary of literature categorized in pretrained models used, tasks tackled and type of transfer learning approach.}
\label{tab:summary}
\endfirsthead
\endhead
\hline
\textbf{Reference}  &  \textbf{Base Model}  &  \textbf{NLP Tasks}     &  \textbf{ Transfer Learning}     \\                  \hline                                     
\citep{mou2016how}         & LSTM-RNN \newline Siamese CNN & Sentiment analysis \newline Sentence pair classification          & Fine Tuning \newline Features Based \\
\hline 
\citep{peters2017semi-supervised} & {CNN-LSTM}                                 & {Named Entity recognition}                                                                                           & Features Based                                                                    \\
\hline 
\citep{lee2017transfer}          & LSTM                                                                   & Named Entity recognition                                                                                                  & Fine Tuning                                                                       \\
\hline
\citep{ruder2017knowledge}     & MLP    & Sentiment analysis & Domain Adaptation              \\
\hline
\citep{mccann2017learned}     & OpenNMT                                                                & {Sentiment analysis \newline Entailment \newline Question Classification \newline Question Answering   }                                                 & Fine Tuning                                                                       \\
\hline

\citep{kim2017cross} & BLSTM             & Part of Speech Tagging & Cross Lingual Fine Tuning                                                                        \\
\hline 
\citep{mccann2017learned}     & OpenNMT                                                                & {Sentiment analysis \newline Entailment \newline Question classification \newline Question answering   }                                                 & Fine Tuning                                                                       \\
\hline 

 \citep{schuster2018cross-lingual}  & ELMoCoVe                                                               & User Intent Classification                                                                                  & Cross-lingual Fine Tuning                                                         \\
\hline 
\citep{lakew2018transfer}    & OpenNMT                                                                & Translation                                                                                               & Fine Tuning \newline Domain Adaptation                                            \\
\hline 
\citep{howardfine-tuned}         & AWD-LSTM                                                               & Text Classification                                                            & Discriminative Fine Tuning                                                        \\
\hline 
\citep{devlin2018bert}             & BERT                                                                   & {Language understanding \newline Natural Language Inference \newline Question Answering}                                                              & {Fine Tuning \newline Features Based}                                                       \\
\hline 
\citep{peters2018deep}             & ELMo                                                                   &{Question Answering \newline Text Entailment \newline Semantic Role Labeling \newline Coreference Resolution \newline Named Entity Extraction \newline Sentiment Analysis}    & Fine Tuning                                                                       \\
\hline 
\citep{shah2018adversarial}             & BLSTM                                                                   &{Duplicate Question Detection}    & Domain Adaptation                                                                       \\
\hline 

\citep{yin2019benchmarking}  & BERT                                                                   &{Topic Categorization \newline Emotion Detection \newline Situation Frame Detection}    & Zero-shot                                                                       \\
\hline 
\citep{stickland2019bert}  & BERT   & Natural Language Understanding  & Adapter Modules                                                                   \\
\hline 
\citep{ma2019domain}     & BERT          &  {Natural Language Inference \newline Answer Sentence Selection \newline Paraphrase Detection} & Domain Adaptation     \\
\hline 
\citep{houlsby2019parameter-efficient}          & BERT                                  & {Text Classification}                                                                        & Adapter Modules                                                                   \\
\hline 
\citep{semnani2019bert}          & BERT                                                                   &Question Answering                                                                        & Adapter Modules \\
\hline 
\citep{radford2019language}            & GPT-2                                                                  & {Language Modeling \newline Children Book Test  \newline Long Range Dependencies  \newline Reading Comprehension \newline Summarization \newline Translation \newline Question Answering} & Zero-shot Transfer                                                                 \\ 
\hline 
\citep{liu2019roberta}              & RoBERTa                                                                & {Question Answering \newline Natural Language Understanding \newline Reading Comprehension }                                                                    & Fine Tuning                                                                       \\
\hline 

\citep{liu2019multi-task}                & BERT                                                                   & Natural Language Understanding                                                                                            & Multi task Fine Tuning                                                            \\
\hline 
\citep{yang2019xlnet}               & XLNet                                                                  & {Natural Language Inference \newline Question Answering \newline Sentiment Analysis \newline Document Ranking}                                         & Fine Tuning                                                                       \\
\hline 
\citep{dong2019unified}  & UNILM & Abstractive Summarization \newline Generative Question Answering \newline Document-grounded Dialog Response  & Fine Tuning \\                                                                      
\hline 
\citep{raffel2019exploring}             & T5                                                                     & {Summarization \newline Question Answering  \newline Text Classification \newline Natural Language Understanding  }                                        & Multi task Fine Tuning                                                            \\
\hline 
\citep{roberts2020how}           & T5                                                                     & Closed Domain Question Answering                                                                                          & Fine Tuning     \\                                                            \hline
\citep{gpt-3} & GPT-3 & Closed Book Question Answering \newline Translation \newline Common Sense Reasoning \newline Reading Comperhension 
\newline Natural Language Understanding \newline Natural Language Inference & Zero-shot \newline Fine Tuning \\
\hline
\end{longtable}

\section{Conclusion}

In this paper, we provided a review of transfer learning in natural language processing. We discussed the possible language models, datasets, and the tasks that were tackled in the research related to transfer learning. Moreover, we provided a taxonomy for transfer learning that divides it into inductive and transductive transfer learning. We then divided each category into multiple levels and then collected the related papers in each corresponding category. Although there might be different definitions in the literature, we tried our best to incorporate the best definition that is agreed upon across different studies. In general, we see that compared to RNN-based and CNN-based language models, it seems that attention-based models are much more dominant in the literature. Additionally, we see that BERT seems the defacto architecture for language modelling as it appears in many tasks. This is due to its bidirectional architectures which makes it successful in many down-stream tasks. Regarding transfer learning, sequential fine-tuning seems to be dominant in the literature as compared to other approaches like zero-shot. Moreover, it seems that mutli-task fine tuning is gaining more attention in the last few years. As stated in many studies, training on multiple tasks at the same time can give much better results.  Regarding datasets, text classification datasets seem to be more widely used compared to other tasks in NLP. This is due to the fact that it is easier to fine-tune models in such tasks. 

For future research, we make some observations and outlooks in the field of transfer learning for NLP. For specific tasks like sentiment classification, abstractive question answering, parts-of-speech tagging, we recommend using bidirectional models like BERT. On the other hand, for generative tasks like summarization, generative question answering, text generation, etc. we recommend using models like GPT-2, T5 and similar architectures. We also believe that some transfer learning techniques are underrated like zero-shot which seems to perform really well on multiple tasks \citep{radford2019language}. Moreover, adapter modules can replace sequential fine-tuning because they perform equally good but provide faster and more compact models as compared to traditional fine tuning. Finally, while language models keep getting bigger and bigger, we believe that more research should be put on trying to reduce the size of such models which will make them deploy-able on embedded devices and on the web. Deploying knowledge distillation \citep{hinton2015distilling} techniques can prove useful in such circumstances for reducing the size of large language models.

\bibliographystyle{plainnat}  
\bibliography{references}

\end{document}